\documentclass[journal]{IEEEtran}
\usepackage{amsmath}
\usepackage{amsfonts}
\usepackage{threeparttable}
\usepackage[pdftex]{graphicx}
\usepackage{hyperref}
\usepackage{booktabs} 
\hypersetup{
    colorlinks=true,
    linkcolor=blue,
    filecolor=blue,      
    urlcolor=blue,
    citecolor=blue,
}

\graphicspath{{images/}}
\DeclareGraphicsExtensions{.pdf,.jpg,.png,.svg}
\begin{document}
\title{IG-RL: Inductive Graph Reinforcement Learning for Massive-Scale Traffic Signal Control}
\author{François-Xavier Devailly, Denis Larocque, Laurent Charlin
\thanks{F-X.\ Devailly, D.\ Larocque, and L.\ Charlin are with the Department of Decision Sciences at HEC~Montr\'eal, Qu\'ebec, Canada. E-mail: francois-xavier.devailly@hec.ca, denis.larocque@hec.ca, laurent.charlin@hec.ca. This work was partially funded by the Natural Sciences and Engineering Research Council (NSERC), Fonds de Recherche du Qu\'ebec: Nature et Technologies (FRQNT), Samsung, and Fondation HEC Montr\'eal. The open source code for IG-RL can be found \href{https://github.com/FXDevailly/IG-RL.git}{https://github.com/FXDevailly/IG-RL.git}.

© 2021 IEEE.  Personal use of this material is permitted.  Permission from IEEE must be obtained for all other uses, in any current or future media, including reprinting/republishing this material for advertising or promotional purposes, creating new collective works, for resale or redistribution to servers or lists, or reuse of any copyrighted component of this work in other works.}}%

\maketitle
\begin{abstract}
Scaling adaptive traffic signal control involves dealing with combinatorial state and action spaces. Multi-agent reinforcement learning attempts to address this challenge by distributing control to specialized agents. However, specialization hinders generalization and transferability, and the computational graphs underlying neural-network architectures---dominating in the multi-agent setting---do not offer the flexibility to handle an arbitrary number of entities which changes both between road networks, and over time as vehicles traverse the network. 
We introduce Inductive Graph Reinforcement Learning (IG-RL) based on graph-convolutional networks which adapts to the structure of any road network, to learn detailed representations of traffic signal controllers and their surroundings. 
Our decentralized approach enables learning of a transferable-adaptive-traffic-signal-control policy. After being trained on an arbitrary set of road networks, our model can generalize to new road networks and traffic distributions, with no additional training and a constant number of parameters, enabling greater scalability compared to prior methods. Furthermore, our approach can exploit the granularity of available data by capturing the (dynamic) demand at both the lane level and the vehicle level. The proposed method is tested on both road networks and traffic settings never experienced during training. We compare IG-RL to multi-agent reinforcement learning and domain-specific baselines. In both synthetic road networks and in a larger experiment involving the control of the 3,971 traffic signals of Manhattan, we show that different instantiations of IG-RL outperform baselines.
\end{abstract}

\begin{IEEEkeywords}
Deep reinforcement learning, Transfer learning, Adaptive traffic signal control, Graph neural networks, Zero-Shot Transfer, Independent Q-Learning
\end{IEEEkeywords}

\section{Introduction}
\IEEEPARstart{T}{he} steady growth of the world's population, combined with a lack of space in urban areas, leads to intractable road congestion, the social and environmental costs of which are well documented~\cite{barth2008real}.
The adaptive control of traffic signal systems based on real-time traffic dynamics could play a key role in alleviating congestion. The framing of adaptive traffic signal control (ATSC) as a Markov decision process (MDP) and the use of reinforcement learning (RL) to solve it via experiencing the traffic system is a promising way to move beyond heuristic assumptions~\cite{chu2019multi,mannion2015parallel, abdulhai2003reinforcement, bingham2001reinforcement}. 

Value-based methods like Q-Learning constitute a cornerstone of RL.
In practice, however, using a single centralized learner~\cite{genders2016using,prashanth2010reinforcement} in an ATSC setting with numerous traffic signal controllers (TSCs) involves a combinatorial action space~\cite{bakker2010traffic}. Multi-agent reinforcement learning (MARL), where each agent controls a single traffic light is appealing, but the proliferation of parameters, nonstationarity, and a lack or transferability (\S~\ref{subsec:decentralization}) make training and scaling challenging~\cite{chu2019multi,mannion2016experimental,wang2019large,kuyer2008multiagent, houli2010multiobjective, mikami1994genetic, prabuchandran2014multi, arel2010reinforcement}. 

\subsection{Contribution}
We introduce inductive graph reinforcement learning (IG-RL), which combines the inductive capabilities of graph-convolutional networks (GCNs)~\cite{hamilton2017inductive} with a new decentralized RL (DEC-RL) framework. In our independent Q-Learning (IQL) formulation, multi-agent RL is replaced by a shared policy, learned and applied in a decentralized fashion. 

We define
the topology of the computational graph of the GCN based on the current dynamic state of the road network (e.g., including the position of moving entities such as cars). 

The GCN learns and exploits representations of a neighborhood of arbitrary order for every entity of the road network, in the form of node embeddings. TSCs' node embeddings are used to evaluate action-values (i.e., Q-values). The entire model is differentiable and we train it using backpropagation of the temporal difference error (TD error), following a standard Deep Q-Learning setting~\cite{szepesvari2010algorithms}.
 By having the computational graph adapt to the road network's state, we are able to:
\begin{itemize}
    \item Train a ubiquitous policy which can adapt to new road networks, including topologies and traffic never experienced during training. 
    \item Exploit the vehicular data at its finest granularity by representing every vehicle as a node and its corresponding vectorized representation (i.e., embedding). Current neural-network architectures used in RL-ATSC do not enable dealing with a changing number of inputs and state dimensionality. For this reason, in a traffic scenario where various types of entities such as cars and pedestrians enter, move inside of, and leave the network, these methods do not enable full granularity exploitation. They  typically resort, upstream of learning, to aggregations at the lane-level (e.g., the number of vehicles approaching the intersection)~\cite{chu2019multi,wei2019colight,wang2019stmarl} or queues lengths~\cite{chu2019multi,wei2019colight,wang2019stmarl}. Alternatively, they fix the number of entities to be represented in detail. For instance, \cite{chu2019multi} represents the cumulative delay of the first vehicle.
\end{itemize}

To test our claims, we define a new evaluation setting in which RL-ATSC methods are tested on road networks which are never experienced during training. In particular, we design two experiments involving non-stationary traffic distributions and different road networks. 

The first experiment, on synthetic road networks, evaluates IG-RL's performance against both learned and domain-specific baselines. We compare a \emph{specialist} instance of IG-RL trained on the road network used during evaluation to a \emph{generalist} instance that is trained on a set of road networks not including the target road network. These instances are compared to assess generalizability, transferability, and the effect of specialization on performance. In addition, we also compare two GCN architectures which respectively capture traffic demand at the vehicle and lane levels, to measure the flexibility of IG-RL in different data conditions, and its ability to exploit granularity.

In the second experiment, an IG-RL instance is transferred, with no additional training (i.e., zero-shot transfer), to the road network of Manhattan and its 3,971 TSCs. This evaluation constitutes, by far, the largest RL-ATSC experiment to date. This final transfer from small synthetic road networks to a large real-world network aims to demonstrate the scalability of learning an agnostic policy. 

\section{Related Work}

\subsection{Decentralization}
\label{subsec:decentralization}
The main thread of research this contribution draws on is decentralized RL (DEC-RL). In RL-ATSC, decentralization and distribution of control via multi-agent RL (MARL) is a popular approach~\cite{chu2019multi,mannion2016experimental,wang2019large,kuyer2008multiagent, houli2010multiobjective, mikami1994genetic, prabuchandran2014multi, arel2010reinforcement}. However, current MARL approaches used for ATSC suffer from the following caveats: 
\begin{itemize}
    \item From the perspective of a given agent (e.g., controller), the evolving behaviors of other agents (e.g., controllers) in the global environment cause nonstationarity and instability during training~\cite{papoudakis2019dealing,omidshafiei2017deep}. As a remedy, researchers include the recent policies of other agents as part of the state of every local agent which seems to help, but does not yet solve the problem~\cite{chu2019multi}.
    \item Without parameter sharing, though  more scalable than previous approaches, computational resources and time required to train MARL approaches grow with every additional signalized intersection as it requires the training of additional agents. 
    \item The specialization of every agent on the particular environment experienced by its jurisdiction during training hinders generalization and transferability to new environments. Therefore, applying MARL on any new road network or after any modification to a road network already under RL control requires training from scratch while interacting with road users and pedestrians to gather experience. 
\end{itemize}

Scalability of MARL and RL-ATSC in general, has therefore typically been limited. In our work, instead of using MARL, we approach DEC-RL as a set of similar decision processes which can be controlled by a single  
agent and show results on a network of 3,971 lights, which is, to the best of our knowledge, the RL-ATSC setting involving the largest number of TSCs to date.

\subsection{Graph-Based Representation}
A second thread which we draw from is graph-neural networks, which have been shown to provide richer spatio-temporal representations of the road network and facilitate coordination~\cite{wei2019colight,wang2019stmarl}. Instead of studying coordination however, we focus on leveraging the flexibility provided by graph-neural networks.

\subsection{Parameter sharing}
Concurrent research~\cite{ wei2019colight} and~\cite{chen2020toward} leverage parameter sharing to enable further scalability. However, our contribution differs significantly from the proposed methodologies in the following ways:
\begin{itemize}
    \item Our main contribution is an inductive method which generalizes and scales to massive networks at test time. Our parameter-sharing scheme enables zero-shot learning---adaptation without additional training---to different road networks. In particular, we demonstrate that our method, trained on small random networks, can control the lights on much larger networks (e.g., with over 3,900 lights) with completely different structures (e.g., traffic lights on the whole of Manhattan).
    \item Our method does not resort to (arbitrary) aggregations of local traffic statistics as state features for parameter sharing. The architecture we use is specifically built to enable full granularity exploitation of demand information sensed at the lane or vehicle levels.
\end{itemize}

\section{Background}
\label{sec:background}

\subsection{Reinforcement Learning (RL)}
\label{subsec:RL}
 The optimization of a decision process under uncertainty can be framed as an MDP. An MDP is defined by a set of states $S$, a set of actions $A$ that the agent can take,  a transition function $\mathcal{T}$ defining the probabilities of transitioning to any state $s^\prime$ given a current state $s$ and an action $a$ taken from $s$, and a reward function $R$ mapping a transition (i.e., $s$, $a$, and $s^\prime$) to a corresponding reward. Optimizing such a decision process means finding a policy $\pi$---a policy is a mapping from a state to the probability of taking an action---which maximizes the sum of the future rewards discounted by a temporal factor $\gamma$:
 \begin{equation}
    \begin{split}
        \begin{gathered}
        R_t=\sum_{k=0}^{\infty}{\gamma^kr_{t+k+1}}
        \end{gathered}
    \end{split}
    \label{equation:rl_reward}
\end{equation}
where $r_t$ represents the reward at time step t. Reinforcement learning (RL) can be used to solve an MDP whose transition and value dynamics are unknown, by learning from experience gathered via interaction with the corresponding environment~\cite{sutton2018reinforcement}. We introduce the key RL concepts that we will use in this paper: 1) Q-Learning a standard algorithm for optimizing a policy in RL problems, 2) the concept of adding parameterized noise for exploration, and 3) decentralized RL. This introduction is self-contained but not exhaustive (an in-depth introduction to the field is available in \cite{sutton2018reinforcement}). 

 \subsubsection{Q-Learning} 
 \label{subsubsec:Q-Learning}
 
 The state-action-value function (Q-function) under policy $\pi$ is the expected value of the sum of the discounted future rewards, starting in state $s$, taking action $a$, and then following policy $\pi$:
\begin{equation}
    \begin{split}
        \begin{gathered}
             {Q}^\pi\left(s,a\right)=\mathbb{E}_\pi\left[\sum_{k=0}^{\infty}{\gamma^k r_{t+k+1}\mid s_t=s,a_t=a}\right]
        \end{gathered}
    \end{split}
    \label{equation:reward}
\end{equation}
From the optimal Q-function $Q^*=\max_{\pi}{Q^\pi}$, we can derive the optimal policy $\pi^\ast\left(a\middle| s\right):a= {argmax}_{a^\prime}Q^\ast\left(s,a^\prime\right)$. $Q^*$ can be obtained by solving the Bellman equation~\cite{bellman1957markovian}:
\begin{equation}
    {Q}\left(s,a\right)=r\left(s,a\right)+\gamma\sum_{s^\prime\epsilon S}\mathcal{T}\left(s^\prime\middle.\mid s,a\right)\max_{a^\prime\epsilon A}{{Q}\left(s^\prime,a^\prime\right)}
    \label{equation:Bellman}
\end{equation}
where $r\left(s,a\right)$ is the immediate reward received after taking action a from state s. Q-Learning uses transitions sampled from the environment and dynamic programming to solve the Bellman equation with no initial knowledge of transition and reward functions. 

There are standard approaches that enable Q-Learning to model environments with large discrete or continuous state spaces (as is the case in this work). One of them consists in fitting the Q-function using buffers of experienced transitions that are large enough to prevent excessive correlations between observed transitions. In continuous state spaces, tabular Q-Learning (shown above) is replaced by a parametric model ${Q}_\theta$ such as linear regression or a deep neural network~\cite{szepesvari2010algorithms}. The latter approach is known as Deep Q-Learning. 

To improve stability a common approach is to keep a lagging version of the model with recent parameters $\theta^-$, called \emph{target model} ${Q}_{\theta^-}$, and use it to compute the temporal difference error (TD error): $r\left(s,a\right)+\gamma\max_{a^\prime}{{Q}_{\theta^-}\left(s^\prime,a^\prime\right)} - {Q}_\theta\left(s,a\right)$. The TD error is backpropagated to update the parameters $\theta$ following the standard Q-Learning update rule: 
\begin{align}
            {Q}_\theta\left(s,a\right)\gets&\;{Q}_\theta\left(s,a\right) \\ &+ \ \alpha\underbrace{\left[r\left(s,a\right)+\gamma\max_{a^\prime\epsilon A}{{Q}_{\theta^-}}\left(s^\prime,a^\prime\right)-{Q}_\theta\left(s,a\right)\right]}_{\text{TD error}} \nonumber
    \label{equation:3}
\end{align}
where $\alpha$ is the learning rate. 

Two standard extensions of Q-Learning led to promising results in our early RL-ATSC experimentation, and we use them in all our experiments: 

\begin{itemize}
    \item Double Q-Learning~\cite{hasselt2010double}, which consists in reducing the overestimation of action values by decoupling action selection and action evaluation.
    \item Dueling Q-networks~\cite{wang2015dueling}, which consists in enabling some parameters to focus on the relative advantage of actions by using additional parameters to evaluate the average of Q-values.
\end{itemize}

  \subsubsection{Noisy Networks for Exploration}
  \label{subsubsec:noisy}
 During training of RL agents, interaction with the environment involves a fundamental dilemma between exploiting what has been learned so far by the agent and gathering additional information~\cite[Chapter~1.1]{sutton2018reinforcement}. Trading off \emph{exploration-exploitation} is critical to learning good policies in a fast and robust fashion, and it remains an active research area. Exploration via noisy networks refers to the use of parameterized noise, in the form of independent Gaussian noise added to the parameters of a neural network, in order to favor consistent and state-dependent exploratory behavior~\cite{fortunato2017noisy}. In comparison to decaying $\varepsilon$-greedy exploration---which consists in decaying the probability of acting randomly during training---noisy-network exploration often leads to better performances, enables online learning since it does not involve limiting exploration as time passes, and does not require tweaking sensitive learning schedules~\cite{fortunato2017noisy}. This form of exploration resulted in considerable improvements during our experiments and we use it to train all models presented in this work.

 \subsubsection{Decentralized Reinforcement Learning}
 Decentralized reinforcement learning (DEC-RL) distributes the control based on the assumption that the global Q-function is decomposable~\cite{busoniu2006decentralized}. Independant Q-Learning (IQL) is a straightforward DEC-RL method in which the global MDP is decomposed into DEC-MDPs which are solved using independent Q-Learners.

\subsection{Graph-Convolutional Networks}
Enabling machine learning model to learn from complex relationships between objects in graphs is challenging. Graph-Convolutional Networks (GCNs) are a recent and fruitful attempt to leverage neural-network architectures and backpropagation to address the complexity of graph data.  GCNs consist in stacking k (a hyperparameter to be defined) convolutional layers as a neighborhood-information aggregation framework~\cite{kipf2016semi}. At every layer of a GCN, every node aggregates \emph{communications} sent to it by both its neighbors and itself into an embedding. This form of embedding both exploits the graph structure and the features of the nodes belonging to its 
neighborhood (up to order k). Using different parameters for different types of relations~\cite{schlichtkrull2018modeling} yields the following message propagation equation, applicable to every node:
\begin{equation}
    \begin{split}
        \begin{gathered}
            n_i^{\left(l\right)}=f\bigg(\sum_{e\epsilon E, j\epsilon N_t\left(i\right)}{C_{i,j,e}\cdot(W_{l_e}\cdot n_j^{\left(l-1\right)}) \bigg)}
        \end{gathered}
    \end{split}
    \label{equation:R-GCN}
\end{equation}
where $n_i^l$ is the embedding of node $i$ at layer $l$, $f$ is a non-linear differentiable function, $E$ is the set of relation types, $N_e(i)$ is the 1st-order neighborhood of node $i$ in the graph of relation type $e$, $C_{i,j,e}$ is a relation-specific normalization constant, and $W_{l_e}$ is the $l^{th}$ layer’s weight matrix for message propagation corresponding to a relation of type $e$. Embeddings can be used to perform a supervised or unsupervised learning task, and an error signal corresponding to the task can be backpropagated through the entire model to perform gradient based optimization of its parameters. 

\section{Inductive Graph Reinforcement Learning (IG-RL)}
\label{sec:IG-RL}

We now introduce IG-RL, a scalable DEC-RL method. IG-RL models objects (e.g., lanes, traffic signals, vehicles) as nodes in a graph. Edges of this (dynamic) graph represent the physical connections between objects (e.g., a vehicle node is connected to its current lane node). 

A GCN models this graph. Initial node embeddings encode observable features of the objects (e.g., the vehicle speed). Then, each inferred TSC-node-embedding is used to predict the Q-values of that TSC. 

The key for generalization is this graph representation. By modeling road networks at the object-level (vehicles, lanes, traffic lights) parameters are independent of any intersection or road-network structure. Therefore, the GCN only instantiates a small number of parameters which are jointly learned across an entire road network or even across multiple road networks. The same parameters can then generalize and transfer, without additional training (zero-shot), to completely different intersections and road networks (e.g., different structures) with different states (e.g., different traffic conditions). 
In turn, this generalization translates into immediate scaling to larger networks independently of the number of intersections.

We first discuss how to decompose the road network into small MDPs and then introduce our GCN model.

\subsection{Decentralized ATSC Decision Processes}
\label{subsec:MDP}
 The state and action spaces of the global ATSC MDP, which consists in the management of all controllers in a given road network, can be large and intractable in practice. We therefore decompose it into smaller decentralized decision processes. In this context, the management of every intersection by a TSC becomes an MDP. 
 
 Each MDP uses continuous states (which represent the state of nearby roads), discrete actions (e.g., whether or not to change phase), and discrete time steps (the length of which determines when the next action can be taken).

\subsubsection{State}
 The state of every decentralized decision process consists in real-time information coming from local sensors and is said to be partially observable because it does not include all information about the global MDP. This state encodes both 1) current connectivity in the network, and 2) demand. Connectivity of a given intersection is defined by the current phase of the corresponding TSC, while demand can be sensed either at the level of lanes, or at the level of vehicles (\S~\ref{subsubsec:Architecture}~and~\ref{subsubsec:features}). 
 
\subsubsection{Action}
\label{subsubsec:action}
 At every intersection of the road network, the flow of traffic is managed by a logical program, composed of a given number of phases, depending on the number of roads, the number of lanes, and the connectivity between lanes. This program cycles through phases, influencing connectivity between lanes, in a constant predefined order that is often known to the road users and pedestrians. In our studies, the agent must respect the program and chooses, every step, whether to switch to the next phase or prolong the current phase. The action space at every intersection is therefore binary. Compared to methods which enable complete freedom of the controller, with the agent being able to choose among any phase every time it picks an action~\cite{chu2019multi}~\cite{prashanth2010reinforcement}, this formulation of the ATSC problem is more constrained. The agent must learn transition dynamics ruled by a complex cycle influencing connectivity. It is however more adaptive than methods determining duration upstream of any given phase~\cite{aslani2017adaptive}. We leave the exploration of other policy constraints for future work, although the model we introduce in \S~\ref{sec:model} is general and is not tailored to the aforementioned constraints. In fact, we discuss modelling an environment where the next action is arbitrary in \S~\ref{subsubsec:action space flexibility}.

\subsubsection{Reward}
The reward for a given agent is defined as the negative sum of local queues lengths $q$, which corresponds to the total number of vehicles stopped\footnote{A vehicle is considered stopped if its speed is inferior to 0.1km/h.} on a lane leading to the corresponding intersection, at a maximum distance of 50 meters of the intersection. The reward at intersection $j$ is
\begin{equation}
    r_j=-\sum_{i\epsilon L_j}{q_i}
    \label{equation:queue_formula}
\end{equation}
where $L_j$ is the set of lanes leading to intersection $j$. This measure can both be obtained via lane or vehicular sensors. We choose this measure instead of alternatives such as wave~\cite{aslani2017adaptive} or average trip waiting time~\cite{mannion2016experimental} because it is correlated to the local transition mechanisms of each DEC-MDP, and it is dense since actions tend to impact this measure quickly. These last two points translate into eased attribution of reward signals~\cite{chu2019multi}.

\subsubsection{Step}
We discretize time into one-second time steps. In other words, actions are selected every second.

\subsubsection{Episode}
Depending on the experiment (\S~\ref{sec:Experiments}), an episode either ends as soon as all trips are completed or after a fixed amount of time.

\subsection{Model}
\label{sec:model}

The control of all traffic signals is a family of decision processes. Each decision process in this family involves a controller, lanes, connections between lanes, and vehicles (e.g., nearby ones). A single GCN is used to model these entities and their dynamic relationships. IG-RL proposes 
systematic sharing of the parameters of the GCN among all objects and relationships of the same nature.

In addition, the flexibility of GCNs enables representing an arbitrary number of entities that changes over time (e.g., vehicles moving in and out of the network) in a detailed way (i.e., as nodes in a graph). 

\subsubsection{Architecture}
\label{subsubsec:Architecture}
\begin{figure}
    \centering
    \includegraphics[width=5cm]{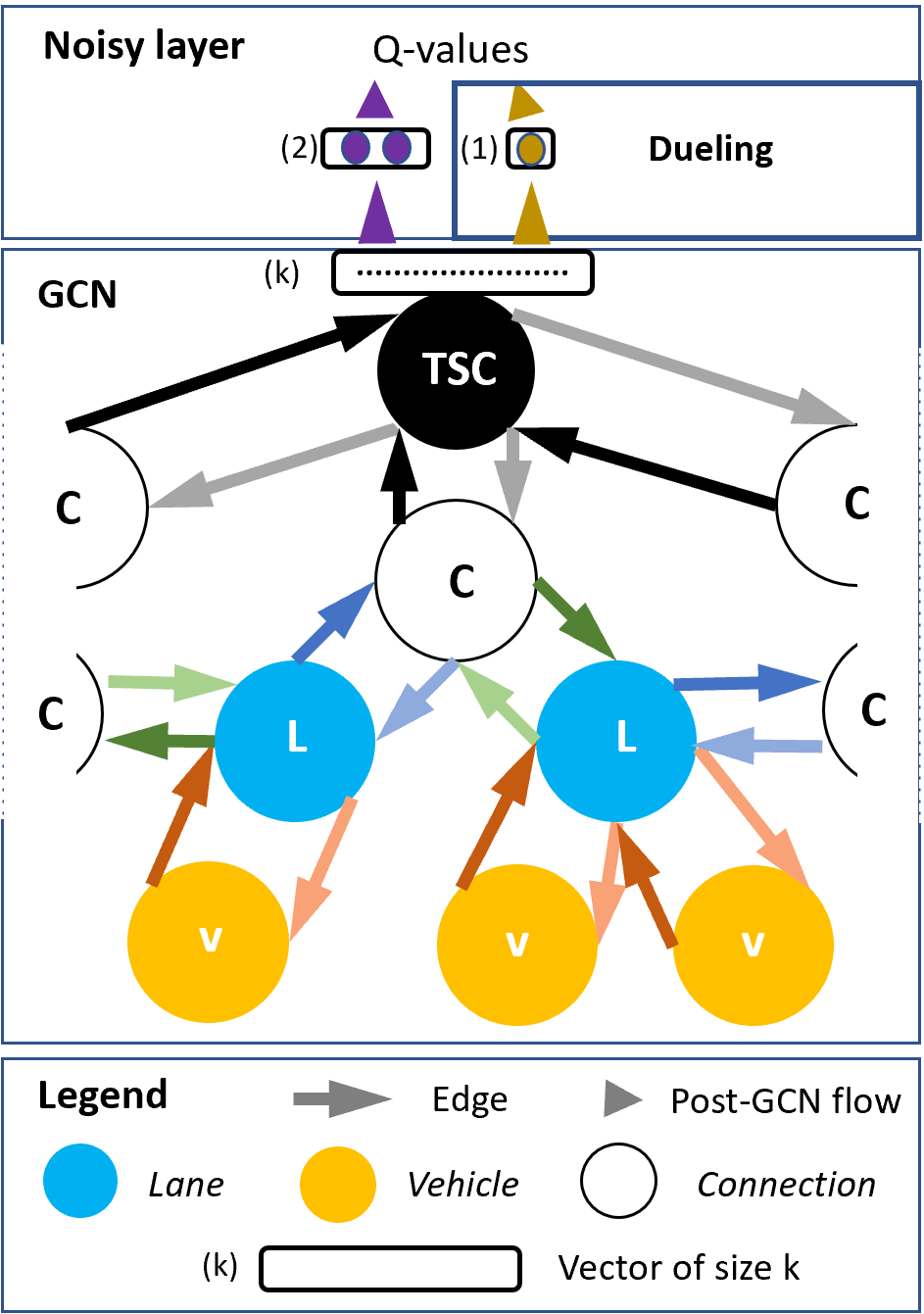}
    \caption{Model. We illustrate the computational graph corresponding to one of the connections a TSC observes at its intersection. One vehicle is located on the connection's inbound lane while two are located on its outbound lane. In this particular example, both lanes are only involved in a single connection on the intersection and respectively start from and lead to other intersections. For every layer of the GCN, message-passing is performed along every edge following Equation~\ref{equation:R-GCN-IG-RL}. Edges of the same type share the same color and parameters. The final embedding of the TSC node is then passed through the a fully connected layer to obtain its Q-values (one per possible action).}
    \label{fig:R-GCN}
\end{figure}

Fig.~\ref{fig:R-GCN} illustrates the GCN for modelling a traffic signal control. It includes 4 types of nodes:
\begin{itemize}
    \item \textit{TSC node}, which represents the state of a controller.
    \item \textit{Connection node}, which represents the state of an existing link between an entry lane and an exit lane.  A link (i.e., connection node) exists between an entry lane A and an exit lane B if, under at least one phase of the TSC program, a vehicle on A is allowed to continue its travel on B. 
    \item \textit{Lane node}, which represents the state of a lane. 
    \item \textit{Vehicle node}, which represents the state of a vehicle.
\end{itemize}

  \noindent The GCN uses the following edges and edge types: 
\begin{itemize}
    \item An edge between every node and itself (4 types of edges).
    \item Bidirectional edges between every \textit{TSC node} and \textit{Connection nodes} corresponding to connections the TSC can influence by changing its phase (2 types of edges).
    \item Bidirectional edges between every \textit{Connection node} and \textit{Lane nodes} corresponding to its entry lane (2 types of edges) and its exit lane (2 types of edges).
    \item Bidirectional edges between every \textit{Lane node} and \textit{Vehicle nodes}  corresponding to vehicles located on the corresponding lane, given that the GCN exploits detailed vehicular data (2 types of links).
\end{itemize}

Every layer of the GCN uses one set of parameters per edge type to perform message-propagation ($W$ in Eq.~\ref{equation:R-GCN}). Our early experimentation with IG-RL and GCNs suggested that using normalization constraints ($C$ in Eq.~\ref{equation:R-GCN}) was detrimental to performance. In contrast to Equation~\ref{equation:R-GCN}, we do not include a normalization constraint during message-passing
\begin{equation}
    \begin{split}
        \begin{gathered}
            n_i^{\left(l\right)}=f\bigg(\sum_{r\epsilon R, j\epsilon N_r\left(i\right)}{W_{l_r}\cdot n_j^{\left(l-1\right)}\bigg).}
        \end{gathered}
    \end{split}
    \label{equation:R-GCN-IG-RL}
\end{equation}
The embedding $n_j^{0}$ of a node $j$ is initialized using its features (\S~\ref{subsubsec:features}).
A fully connected layer, with a single set of parameters, maps the final node embedding of a TSC to the Q-values corresponding to its selectable actions. 

Since preliminary experiments demonstrate that noisy exploration enables faster and more robust training, as well as better performance compared to $\varepsilon$-greedy exploration, parameterized Gaussian noise is added to the parameters of this final mapping. 
An initialization of 0.017 for the variance parameters performs well in multiple supervised~\cite{fortunato2017noisy} and reinforcement learning tasks~\cite{fortunato2017bayesian}, and so we also use that value.
Fig.~\ref{fig:R-GCN} is an illustration of the architecture of the full model for a simple road network with a single TSC and three vehicles.

\subsubsection{Action Space Flexibility}\label{subsubsec:action space flexibility}
Given the computational flexibility of GCNs, the model can be easily adapted to different action spaces. For instance, 
enabling the choice of an arbitrary phase (instead of a program that cycles through phases) is achieved by representing phases as nodes of a new type, and linking these nodes to both their corresponding TSC node and the nodes representing connections they influence. This way, one Q-value can be computed per phase node. Preliminary experiments using this formulation seem promising.

\subsubsection{Parameter Sharing}\label{subsubsec:parameter sharing}
Using a graph representation at the object level (TSC, connection, lane, car) enables sharing parameters between objects of the same type, both inside each decision process (e.g., two lanes on the same intersection), and across all decision processes belonging to the same family (e.g., two lanes on unrelated intersections). This parameter sharing is useful because: 
\begin{itemize}
    \item It uses a constant number of parameters independent of the number of underlying processes.
    \item During training, these shared parameters are updated using experience gathered from a variety of decision processes (a variety of intersection architectures belonging to a variety of road network architectures).
\end{itemize}
    
\subsubsection{Generalization and Transfer via Inductive Learning}\label{subsubsec:Generalization and Transfer via Inductive Learning} 
Consistent parameter sharing enables training a unique set of parameters using experience gathered from a set of decision processes involving diverse intersection topologies, road network topologies, and traffic distributions. Enforcing such diversity can translate into better generalization to unseen MDPs from the same family of MDPs~\cite{sadeghi2018sim2real}~\cite{akkaya2019solving}. In this work, the learning of general and transferable patterns is based on two key components: 
\begin{itemize}
    \item The proposed architecture, based on a GCN, which enables parameter sharing without sacrificing granularity.
    \item The sampling strategy over the family of MDPs, which in this context refers to the way decision processes are created. The sampling strategy should ensure that the training set is representative of the corresponding family of decision processes.
\end{itemize}

\subsubsection{Marginalizing Nonstationarity}\label{ subsubsec:Marginalizing nonstationarity}   
Nonstationarity, caused by the evolution of the behavior of neighbor agents during training, remains a critical challenge for DEC-RL (\S~\ref{subsec:decentralization}).
We hypothesize that emphasizing diversity in the training set, as described in \S~\ref{subsubsec:parameter sharing}, can limit nonstationarity and stabilize learning by marginalizing out the experience of a particular intersection with respect to its neighbor intersections.

\subsubsection{Features}\label{subsubsec:features}
Our approach exploits data that can be accessed from sensors representing the various entities that make up the road network and traffic. At any time step in a given road network, every element defining the state is represented through a node in the GCN. The architecture of the GCN is built on the assumption that the current state of the controller (i.e., the current phase and its corresponding local connectivity) is known, and that traffic information coming either from lane sensors or vehicle sensors is also available. Other assumptions would result in the choice of a different GCN architecture or in the use of different features. The features used for different types of nodes are summarized in Table~\ref{table:nodes_variables}. \textit{Current speed} represents the current speed of a vehicle in km/h, \textit{Position on lane} represents the relative location of a vehicle with respect to the lane it is on. It is defined as a proportion of the lane’s length. \textit{Length} is the length of a lane in meters, \textit{Number of vehicles} is the total number of vehicles captured by the lane’s sensor, \textit{Average vehicles speed} is the average speed of vehicles captured by the lane’s sensor, \textit{Is open} represents whether a connection is opened under the current phase, \textit{Has priority} represents whether, if an open connection between an entry and an exit lane has priority\footnote{Vehicles may use a connection (i.e., go from its entry lane to its exit lane) if no vehicle uses a higher prioritised connection} or not. \textit{Number of switches} to open is the number of switches the controller has to perform before the next opening of a given connection. \textit{Next open has priority} defines whether the next opening of the connection will have priority or not. \textit{Time since last switch} is the number of seconds since a traffic controller performed its last phase switch.
\begin{table}[h!]
\centering
\begin{threeparttable}
\caption{Initial state variables for every type of node.}
\label{table:nodes_variables}
\begin{tabular}{cl}
\toprule
\textbf{Type of node}  & \textbf{State features}                                                                                                    \\ \midrule
{Vehicle}      & \textit{Current} speed, \textit{Position on lane}                                                                                       \\ \midrule
{Lane}          & \textit{Length}, \textit{Number of vehicles}, \textit{Average vehicles speed}                                                                     \\ \midrule
{Connection}    & \begin{tabular}[l]{@{}l@{}}\textit{Is open}, \textit{Has priority}, \textit{Number of switches to open},\\  \textit{Next opening has priority}\end{tabular} \\ \midrule
{TSC} & \textit{Time since last switch}                                                                                                \\ \bottomrule
\end{tabular}

\end{threeparttable}
\end{table}

\section{Experiments}
\label{sec:Experiments}

Using the SUMO traffic simulator~\cite{SUMO2018}, we evaluate the performance of several IG-RL instantiations using small synthetic road networks and a large real road network, and compare it to the results of several baselines. 

Our study demonstrates that: 
\begin{itemize}
    \item IG-RL outperforms all included baselines on the ATSC task in a traditional evaluation setting.
    \item Training IG-RL under several varying initial conditions (including different road networks) yields the best performing method and enables zero-shot transfer.
    \item Training IG-RL on small synthetic road networks with limited computational resources enables efficient zero-shot transfer to real road networks, and such a transfer translates into improved scalability. 
\end{itemize}

The design of a synthetic training set (of random road networks) before transferring the learned policy to a more realistic setting (Manhattan) is inspired by other zero-shot transfer approaches in reinforcement learning~\cite{oh2017zero}~\cite{higgins2017darla}.

\subsection{General Setup}

 \subsubsection{Network Generation}
 \label{subsubsec:network generation}
\begin{figure}
    \centering
    \includegraphics[width=8cm]{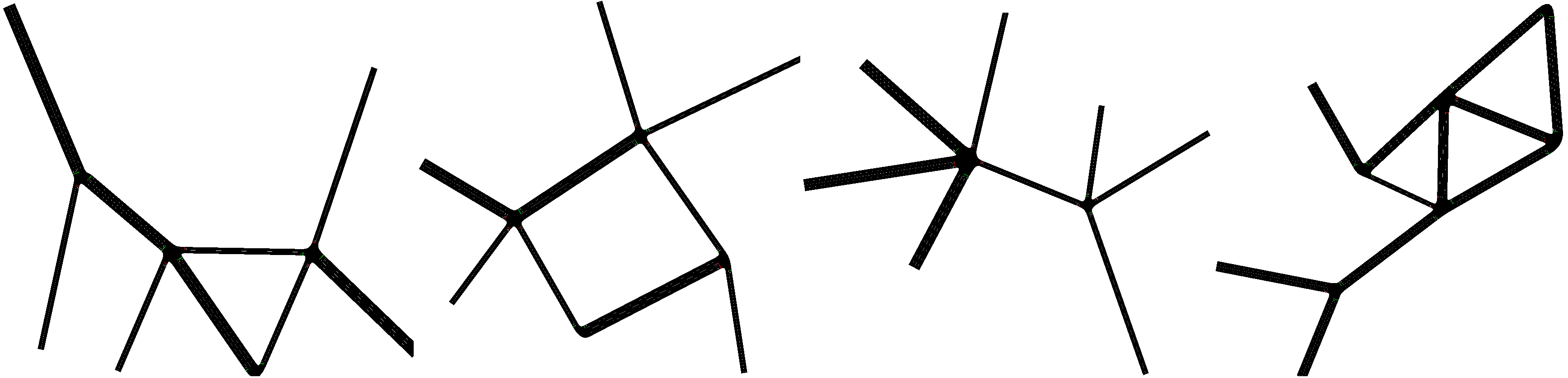}
    \caption{Four randomly generated road networks. Thickness indicates the number of lanes per direction (between 1 and 2 per direction for a maximum of 4 lanes per edge).
    }
    \label{fig:networks}
\end{figure}

The structure of a road network is picked, at random, from a set which consists in the cross product of the number of intersections (typically between 2 and 6), the structure of every intersection, the length of every edge (between 100 and 200 meters), the number of lanes per route (between 1 and 4 per edge), and general connectivity. A given random seed will lead to the generation of the same network, enabling exact comparison of performance between policies. Examples of randomly generated networks from our procedure are shown in Fig.~\ref{fig:networks}.

\subsubsection{TSC-Programs} 
TSC-programs, or cycles, are generated by SUMO and typically include several phases, corresponding to the legal states at a given intersection.\footnote{``All red'' phases are not included as part of SUMO's default TSC-Programs generation.} Programs are therefore not restricted to abstracted axis connectivity such as North-South or East-West. Our synthetic network generation procedure~(\S~\ref{subsubsec:network generation}) leads, on average, to the creation of TSCs whose programs are composed of 5.5 phases. We leave experimentation with other types of TSC-programs to future work. 

\subsubsection{Traffic Generation}
\label{subsubsec:Traffic Generation}
A given number of trips are to be generated per second, on average, during an episode. This value is referred to as the traffic regime. Every generated trip is assigned a trajectory. A trajectory is defined by a starting lane, a sequence of intermediary lanes, and a final lane. Every valid trajectory includes at least 2 lanes. To simulate the urban setting, a trajectory can start and end anywhere in the network. Parameters defining non-uniform distributions used for assigning trajectories to trips are re-sampled every 2-minutes to ensure non-stationary distributions.
\subsubsection{Training}
\label{subsubsec:training_curriculums}
Models are trained on a set of 30 simulations running in parallel. During training, exploratory behaviors tends to lead to catastrophic congestion from which recovery becomes unlikely. This is a common RL challenge which makes training from long episodes inefficient in practice. For this reason, as in~\cite{chu2019multi}, every simulation runs 30,000 steps divided into independent short episodes of 500 steps (seconds) each. All simulations run on a single randomly generated road network. We also use this network for evaluation in our first experiment (\S~\ref{sec:experiment1}). Further, each simulation is initialized using a different random seed to ensure a variety of observed transitions. Models are trained using the exact same environment (network and traffic). The parameters of target Q-networks are updated every 100 parameters updates of the main Q-networks (\S~\ref{subsubsec:Q-Learning}) The learning rate is 0.001 and the batch size is 16.\footnote{A batch is composed of networks. A network typically involves multiple intersections}
    
 \subsubsection{Evaluation}
 \label{subsubsec:evaluation}
All methods are evaluated using the same target networks and identical traffic. 
However, evaluation introduces traffic regimes (densities) never experienced during training, to evaluate how the methods generalize to new traffic distributions. 
We evaluate performance by measuring the evolution of the instantaneous delay:
\begin{equation}
    \begin{split}
        \begin{gathered}
            s_{v}^{*}=min(s_{v^*},s_l), \\
            d_t=\sum_{v\epsilon V}(s_{v}^{*}-s_v)/s_{v}^{*}
        \end{gathered}
    \end{split}
    \label{equation:delay}
\end{equation}
where $V$ is the set of all vehicles in the road network, $s_{v^*}$ is the maximum speed of the vehicle, $s_l$ is the maximum allowed speed on the lane the vehicle is on, $s_v$ is the current vehicle speed, and $d_t$ is the total instantaneous delay at time step $t$. Given that all models are evaluated on the same sets of networks and trips, we can pair the same trips together for hypothesis testing (using paired t-tests), and to study distributions of trips-durations differences.  
 \subsubsection{Constraints}
 \label{subsubsec:constraints}
Phases involving yellow lights last exactly five seconds, as suggested by~\cite{aslani2017adaptive}. For all other phases, in order to maximize adaptiveness, instead of using a fixed time of 5 seconds between every action as suggested by~\cite{chu2019multi}, 
a TSC must spend at least 5 seconds in each phase before being allowed to switch.
The parameters associated with these time constraints can be modified and their values are expected to influence achievable performance. We leave thorough
experimentation with time constraints (e.g., minimum green time for pedestrians) to future work.
For trainable methods, constraints are enforced by ignoring invalid actions selected by the agent. In other words, in a given time step, even though the agent can pick an action, its decision will  have no influence on the MDP if its chosen action breaks a constraint.

\subsubsection{Action-Correction}
\label{subsubsec:action-correction}
Ignoring an agent's invalid actions forces it to learn that the effect of the available actions on connectivity are dependent on other logical factors like the time since the last phase change and the nature of the current phase. To ease learning, and since it is preferable to focus on the actual influence an agent can have on its MDP, we replace, during experience replay, binary decisions which were chosen by the agent at every time step, by whether a light switch was performed. We also force the greedy policy, in the creation of the temporal difference target, to only consider actions which could be performed at the corresponding step. To adapt the proposed action-correction methodology to a policy which is able to choose an arbitrary phase, the selected action can simply be replaced by the action corresponding to the selection of the phase that is actually active at the next step.

\subsubsection{Robustness}
The seeds we use in the experiment influence the random generation of  road networks, traffic distributions, initial weights of neural networks, and the Gaussian noise used for exploration during training. To ensure the robustness of our conclusions, every experiment, including both training and test, is repeated 5 times using different seeds. For instance, the training on 30 different simulations occurs 5 times for a total of 150 simulations. The same goes for evaluation. Reported test results are an aggregation over these 5 runs for both experiments.
\subsection{IG-RL models}

\subsubsection{Architectures}
\label{subsubsec:architectures}
Our complete GCN architecture, referred to as IG-RL-Vehicle (IG-RL-V), uses a 3-layer GCN and vehicle-level traffic information for both state\footnote{Lane level traffic features are not included in IG-RL-V.} and reward (\S~\ref{subsec:MDP}). 

We also introduce a lighter GCN architecture, IG-RL-Lane (IG-RL-L) which uses a 2-layer GCN and lane-level traffic information for both state\footnote{Vehicle nodes are not included in IG-RL-L.} and reward (\S~\ref{subsec:MDP}). Even though IG-RL-L does not exploit demand at the vehicle-level, it has the added benefit of requiring constant computational time with respect to the traffic demand. 

For both IG-RL-L and IG-RL-V, the number of layers is chosen to be the minimum which ensures that information representing demand can reach the nearest TSC(s) nodes. For both GCN architectures, we use node embeddings of size 32 as further increasing this hyperparameter did not result in improved performances.  

\subsubsection{General versus Specialized}
The target network refers to the road network used to test the performance of all methods after training and must be distinguished from the training set of road networks since the same IG-RL instance can be applied, and therefore trained on, different road networks.
Two training sets of road networks are built for IG-RL. 

The \emph{specialist set}, used to train Specialist-IG-RL (S-IG-RL), aims to evaluate intentional specialization on a given network. It consists in training the model exclusively on the target road network, as described in \S~\ref{subsubsec:training_curriculums}. 

The \emph{generalist set}, used to train Generalist-IG-RL (G-IG-RL), aims to study IG-RL's transfer abilities. This set is composed of independent road networks and does not include the target network. In such a context, performance reflects generalizability to new road network and intersection architectures. In addition, as different road networks imply different traffic distributions even when the regime (\S~\ref{subsubsec:evaluation} and \S~\ref{subsubsec:description exp1}) is left unchanged, performance reflects generalizability to new traffic distributions for all studied regimes, even the \emph{default} one which is used during training.

Our experiments with hybrid training sets involving both the target road network and random road networks did not lead to improved performance compared to the generalist approach and are omitted for brevity.

\subsubsection{Evaluating Action-Correction}
To evaluate the importance of action-correction (\S~\ref{subsubsec:action-correction}), we introduce a version of IG-RL-L for which no action-correction is performed during training. This ablation study is referred to as IG-RL-no-correction (IG-RL-n-c).

\subsection{Baselines}
\label{subsec:Baselines}
We now describe the models used as baselines in both of our experiments. 

 \subsubsection{Fixed-Time Baseline}
This baseline follows the cycle of phases using predefined and constant durations generated by SUMO~\cite{SUMO2018} based on the architecture of each intersection. 

 \subsubsection{Max-moving-car heuristic (Greedy)}
This dynamic heuristic-based method aims at ensuring that as many vehicles as possible are moving on inbound lanes at any given time. At every intersection, the controller switches to the next phase if, on inbound lanes, the number of stopped vehicles is superior to the number of moving vehicles, and prolongs the current phase otherwise.

 \subsubsection{MARL-IQL}
 \label{subsubsec:MARL-IQL Details}
MARL-IQL is a learned baseline. In this model, every intersection is controlled by its corresponding agent (i.e., with its own unique parameters), which learns to solve its own local MDP. Every agent, parameterized by a deep Q-Learner network, takes the exact same variables that are available to IG-RL-L as inputs (Table~\ref{table:nodes_variables} and \S~\ref{sec:Experiments}), with the exception of lanes’ \textit{lengths}, which is constant for each MDP. Every Q-Learner consists of a first hidden layer of 256 neurons, a second hidden layer of 128 neurons, and a last hidden layer of 64 neurons, which is the result of a grid-hyperparameter search. While this model would constitute a ``vanilla'' multi-agent implementation, we chose to add both Double Q-Learning and a Dueling architecture to every Q-Learner (\S~\ref{subsubsec:Q-Learning}) as done in~\cite{liang2019deep}. In addition, the introduction of noisy parameters (\S~\ref{subsubsec:noisy}) and action-correction (\S~\ref{subsubsec:action-correction}), were the elements which led the the biggest improvements during training. 
These improvements have the two following objectives: 
\begin{itemize}
    \item Create a strong baseline.
    \item Enable an ablation study. The only difference between MARL-IQL and S-IG-RL-L is that the former uses one neural network per intersection to learn the local Q-function and compute the local Q-values, while the latter uses the same GCN to do so at every intersection. Apart from this, all the information and methods they use for training are identical.
\end{itemize}

\subsection{Experiment 1: Evaluating Generalization using Synthetic Road Networks }\label{sec:experiment1}
\subsubsection{Description}
\label{subsubsec:description exp1}

In this experiment, we evaluate policies on a single road network. The same road network is used to train all methods except G-IG-RL. As MARL-IQL does not scale as efficiently as IG-RL methods, we picked a small road network.  %

\paragraph{Evaluation setting}
During evaluation, trips are generated during the first hour. An episode ends as soon as all trips are completed. Performance of all methods are evaluated using 30 different random seeds for traffic generation. Evaluation is run twice, with two distinct traffic regimes (denoted \emph{default} and \emph{heavy}), in order to evaluate the ability of models to generalize to traffic densities never experienced in training, as well as to study the performance in different density settings. The first evaluation traffic regime, which is the one used during training, introduces an expected one vehicle every 4 seconds in the network, while the second traffic regime introduces twice as many per second on average.

\paragraph{Overcoming overfitting}
\label{paragraph:overcoming overfitting}
While methods are evaluated during episodes of more than an hour, they are trained on shorter episodes of 500 seconds (\S~\ref{subsubsec:training_curriculums}). As  more  vehicles  enter  the  network, congestion can therefore increase past levels experienced during training (which again implies new traffic distributions) even in the \emph{default} traffic regime used during training. Since conditions differ between training and evaluation, the task is very challenging and the learning of generalizable patterns and behaviors (policies) is key to perform in both traffic regimes.

 \subsubsection{Results}

\begin{figure}[htp]
    \centering
    \includegraphics[width=9cm]{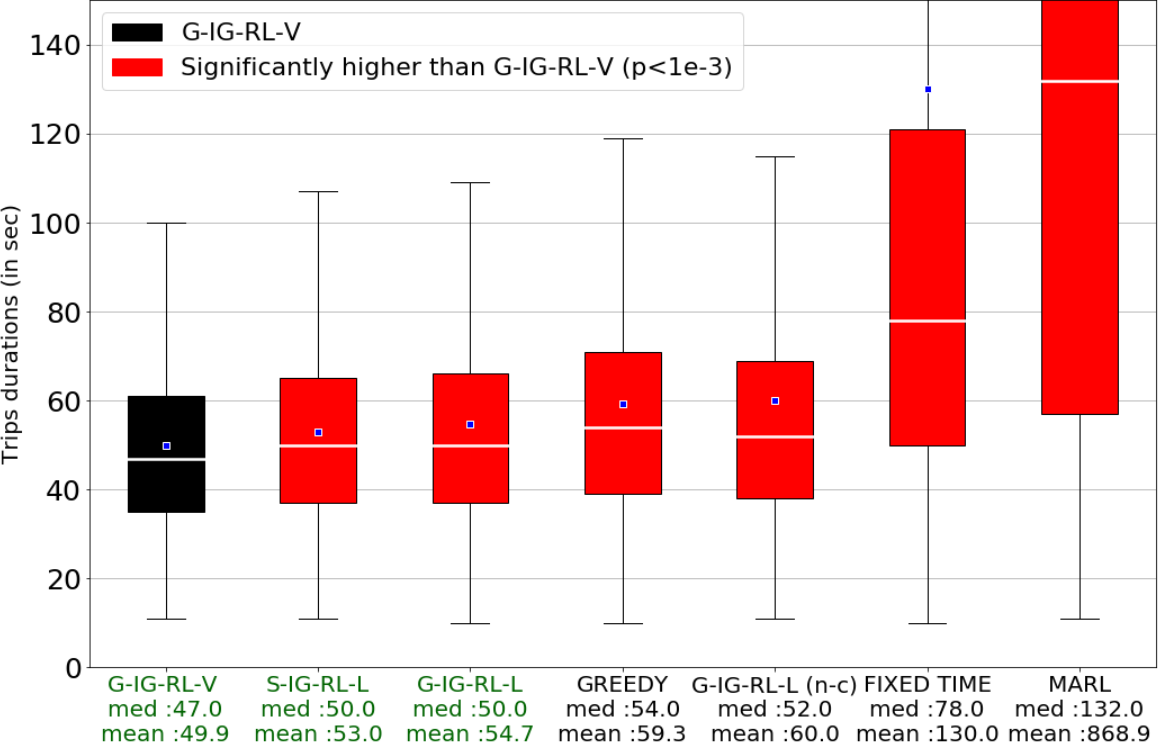}
    \caption{Trips Durations: \textbf{Default} Traffic Regime | Synthetic Road Networks.\\\hspace{\textwidth}This figure presents the distribution of trips durations during test in the same traffic regime used for training. The vertical axis is cut at MARL's median trip duration for readability. Since every method is evaluated using the exact same trips, for every method, we perform t-tests on the paired sample of: 1) The durations of the trips when using the given method  and 2) The durations of the trips when using G-IG-RL-V. The results of these paired t-tests suggest that G-IG-RL-V significantly outperforms every other method, while the boxplots, means, and medians of trips durations suggest that IG-RL models using action-correction outperform all other methods in general.}
    \label{fig:trips_durations_L}
    
    \bigskip
    \centering
    \includegraphics[width=9cm]{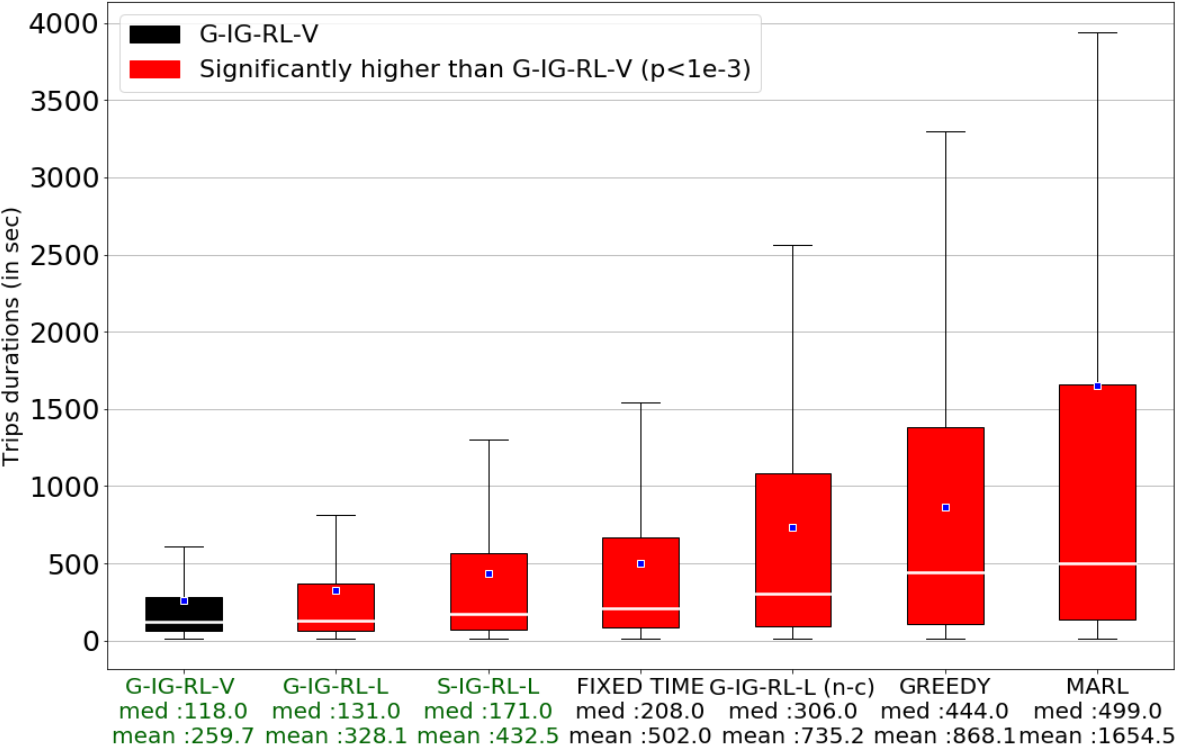}
    \caption{Trips Duration: \textbf{Heavy} Traffic Regime | Synthetic Road Networks.\\\hspace{\textwidth}This figure presents the distribution of trips durations during test in the heavier traffic regime that was not experienced during training. Results are again in favor of G-IG-RL-V and IG-RL methods in general. Performance gains are larger in this setting compared to the \emph{default} traffic regime. (Fig.~\ref{fig:trips_durations_L})}
    \label{fig:trips_durations_H}    

\end{figure}
\begin{figure}[htp]
    \centering
    \includegraphics[width=9cm]{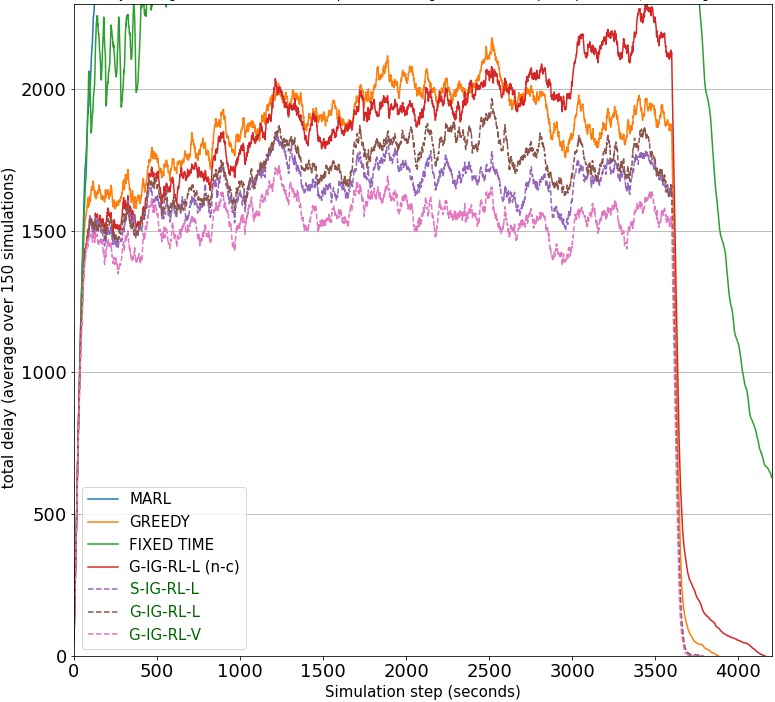}
    \caption{Total Delay Evolution: \textbf{Default} Traffic Regime | Synthetic Road Networks. For clarity, this figure focuses on competitive approaches with lower delays (which stabilize early on). The delays for Fixed-Time and MARL continue to increase and do not converge within an hour. A complete version of this figure is provided as supplementary material (\S~\ref{sec:supplementary_material}).}
    \label{fig:step_delay_L}
    \bigskip
    \centering
    \includegraphics[width=9cm]{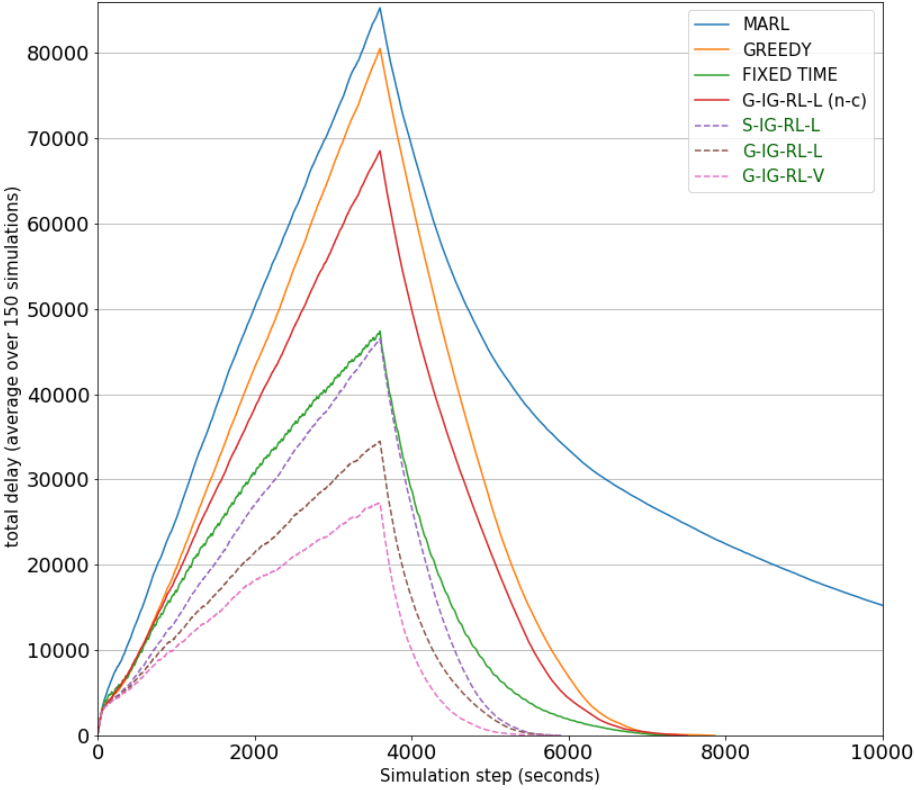}
    \caption{Total Delay Evolution: \textbf{Heavy} Traffic Regime | Synthetic Road Networks.}
    \label{fig:step_delay_H}
    
\end{figure}

\begin{figure}[htp]
    \centering
    \includegraphics[width=9cm]{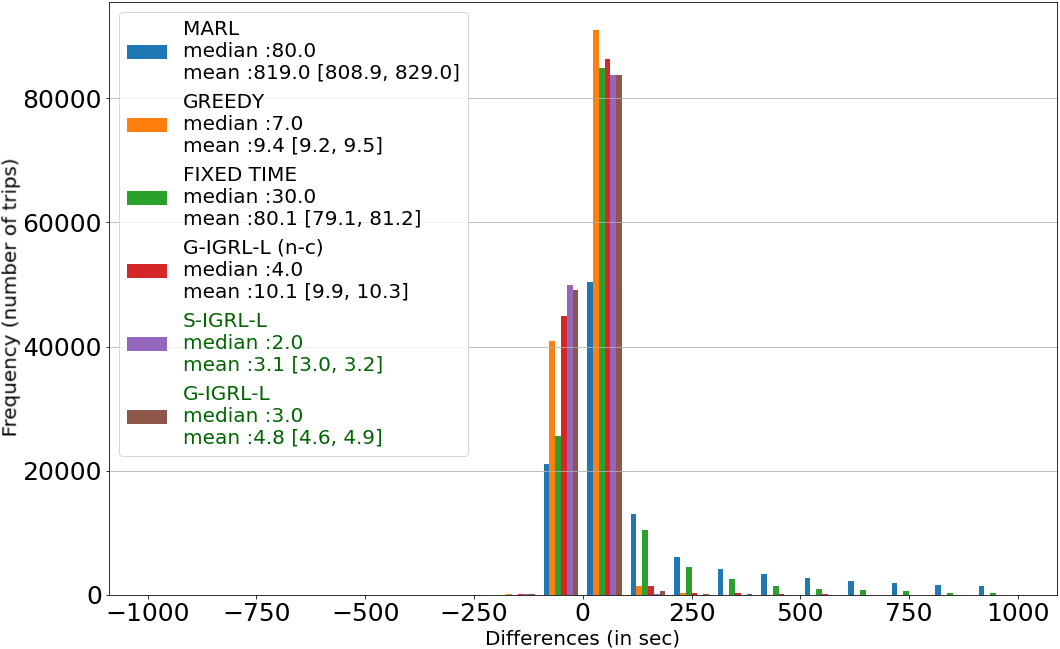}
    \caption{Differences of Paired Trips Durations (compared to G-IG-RL-V): \textbf{Default} Traffic Regime | Synthetic Road Networks. This figure presents the histogram of the differences of paired trips durations (\S~\ref{subsubsec:evaluation}) between G-IG-RL-V and each other method. Evaluation involved the same traffic regime as used in training. Positive values indicate trips which were completed faster using G-IG-RL-V than the compared method. Negative values indicate trips which were completed faster using the compared method than G-IG-RL-V. Medians, means, and confidence interval for means of the differences are also reported. All these results show that when using G-IG-RL-V, trips tend to be completed faster .}
    \label{fig:diff_distrib_L}
    \bigskip
    \centering
    \includegraphics[width=9cm]{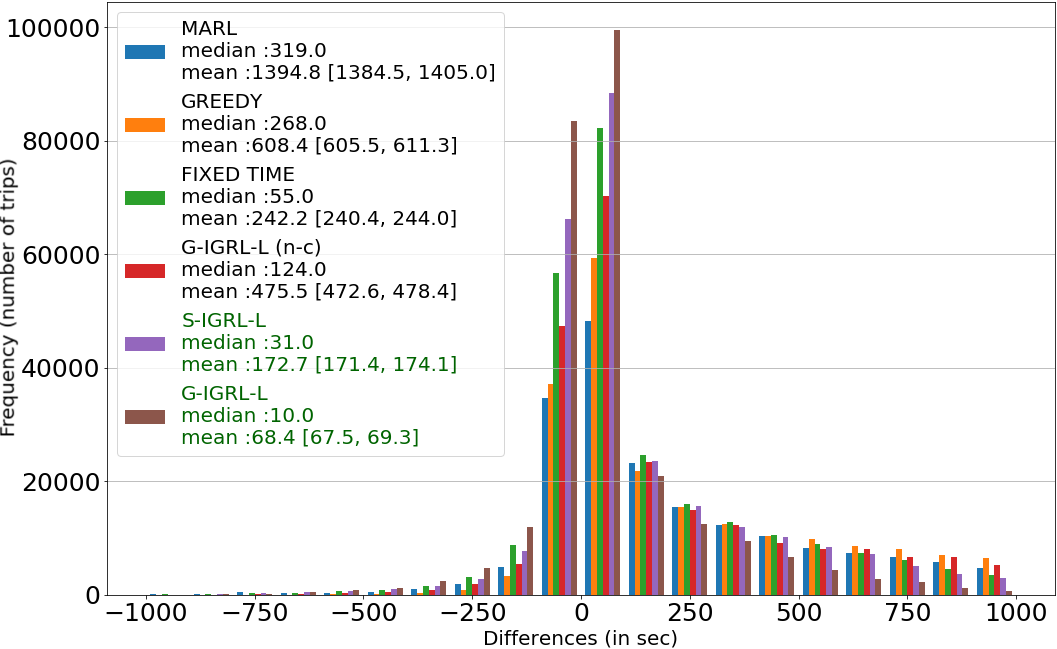}
    \caption{Differences of Paired Trips Durations (compared to G-IG-RL-V): \textbf{Heavy} Traffic Regime | Synthetic Road Networks. This figure reports the same metrics as Fig.~\ref{fig:diff_distrib_L}, but the evaluation uses the heavier traffic regime not experienced during training. Results are again in favor of G-IG-RL-V, with larger differences against all compared methods than under the \emph{default} traffic regime. In addition, extreme delays are avoided when using G-IG-RL-V.} 
    \label{fig:diff_distrib_H}
    
\end{figure}

\paragraph{IG-RL models}
We now refer to the group of methods using both IG-RL and action-correction as IG-RL models, and highlight them in all following figures (green font).
Means and medians of trip durations, reported in Figs.~\ref{fig:trips_durations_L}~and~\ref{fig:trips_durations_H}, are lower when using IG-RL models, compared to other methods. Instantaneous total delays, reported for every simulation step in Figs.~\ref{fig:step_delay_L}~and~\ref{fig:step_delay_H}, show that congestion increase is slower, and recovery\footnote{Recovery refers to the speed at which delay decreases after the simulation reached 1 hour and no additional trips are generated.}  is faster using IG-RL models in both traffic regimes. In addition, in the \emph{default} traffic regime, these methods reach the lowest steady-state in terms of congestion level.

G-IG-RL-V is the best performing method, with both the lowest trip-durations (Figs.~\ref{fig:trips_durations_L}~and~\ref{fig:trips_durations_H}), the slowest congestion increase and the fastest recovery (Figs.~\ref{fig:step_delay_L}~and~\ref{fig:step_delay_H}). In addition, distributions of paired-trips-durations differences, between every method and G-IG-RL-V, reported in Fig.~\ref{fig:diff_distrib_L}~and~\ref{fig:diff_distrib_H}, show that a large majority of trips are completed faster using G-IG-RL-V. In the \emph{default} traffic regime, some trips are delayed by up to a thousand seconds when using MARL or Fixed-Time. In comparison, no trips are delayed by more than a hundred seconds under the G-IG-RL-V policy. In the \emph{heavy} traffic regime, the conclusions are similar, with even higher  differences compared to other methods. Fig.~\ref{fig:diff_distrib_L}~and~\ref{fig:diff_distrib_H} also show that in addition to improving average performance, G-IG-RL-V tends to distribute delay in a more equitable way than any other method (numerous extreme delays are avoided and replaced by a few shorter delays).

\paragraph{G-IG-RL-L and S-IG-RL-L} A key finding emerges from comparing G-IG-RL-L and S-IG-RL-L. Even though their performance is similar in the \emph{default} regime, G-IG-RL-L outperforms S-IG-RL-L in the \emph{heavy} regime.
This shows that G-IG-RL enables the efficient learning of transferable patterns, and that learning from a variety of road networks is key to enable generalization to new traffic regimes. This result further emphasizes the limits of training any policy, whether it is a single policy in the context of centralized RL or multiple policies in the context of MARL, on a single road architecture. (Recall that allowing for modelling different sets of train and test networks is one of our key contributions.)

\paragraph{IG-RL-V and IG-RL-L} 
G-IG-RL-V outperforming other methods in both traffic regimes highlights the flexibility of the GCN to exploit a finer level of granularity (i.e., data sensed at the vehicle level). However, G-IG-RL-L and S-IG-RL-L being the second bests (ex-aequo) performing methods in the \emph{default} traffic regime, and G-IG-RL-L being the second best performing method when generalizing to a heavier traffic regime, we can conclude that IG-RL can also perform well with demand sensed at the lane level. As the comparison between S-IG-RL-V and S-IG-RL-L leads to the same conclusion as the one between G-IG-RL-V and G-IG-RL-L, we did not include S-IG-RL-V in this paper to improve readability.

\paragraph{Action-Correction} The performance of G-IG-RL-L(n-c) shows that removing action-correction (\S~\ref{subsubsec:action-correction}) has an important negative impact on performance and generalization to the \emph{heavy} traffic regime, with trips lasting longer, congestion increasing faster, and recovering slower than every other IG-RL-based method.

\paragraph{MARL-IQL}
\label{subsubsec:IG-RL Vs. MARL-IQL}
MARL-IQL is the worst performing model (it is even outperformed by the fixed-time baseline).
We hypothesize that two reasons explain its performance:

\begin{itemize}

    \item Overfitting: As mentioned in~\S~\ref{paragraph:overcoming overfitting}, the current experiment is very challenging with respect to overfitting in both traffic regimes. During training, we observed that MARL-IQL largely outperforms the Fixed-Time policy, and its performance is both: 1) on par with the performance of the Max-moving-car heuristic (Greedy) policy, and 2) much closer to the performances of other learned models as compared to the gap in performance at test time.
    This observation suggests that policies learned by MARL-IQL do not generalize well to new demand and traffic distributions. 
    \item Nonstationarity: As described in~\S~\ref{subsec:decentralization}, nonstationarity is a key reason for instability in MARL. It is no surprise that naively applying deep neural networks to MARL-IQL can lead to subpar performance. Note that this is the case even though we use an improved version MARL-IQL (\S~\ref{subsec:Baselines}). In \cite{chu2019multi}, for instance, a MARL-IQL method performed significantly worse than all other formulations, including one based on vanilla linear regression.

\end{itemize}

\paragraph{Conclusions}
This experiment highlights that IG-RL's 1) parameter-sharing (\S~\ref{subsubsec:parameter sharing}), 2) training schemes (\S~\ref{subsubsec:Generalization and Transfer via Inductive Learning}), and 3) vehicle-level-demand representation (\S~\ref{subsubsec:architectures}) are key to its performance. The first is observed when comparing MARL-IQL to IG-RL models (parameter sharing distinguishes S-IG-RL-L from MARL-IQL as stated in  \S~\ref{subsubsec:MARL-IQL Details}). The second is shown by the performance of the G-IG-RL models (especially in the \emph{heavy} traffic regime). Further, training all parameters with a variety of experiences (from multiple intersections using S-IG-RL and even multiple road networks using G-IG-RL) and avoiding overparameterization helps IG-RL prevent overfitting (a key challenge in this experiment as stated in~\S~\ref{paragraph:overcoming overfitting}). We hypothesize that this variety also marginalizes out nonstationarity (\S~\ref{ subsubsec:Marginalizing nonstationarity}). 
Finally, comparison of G-IG-RL-V and G-IG-RL-L demonstrates the third highlight.

\subsection{Experiment 2: Transfer and Scaling to Manhattan}
 \subsubsection{Description}
 \label{subsubsec:Exp2 description}
\begin{figure}[htp]
    \centering
    \includegraphics[width=7cm]{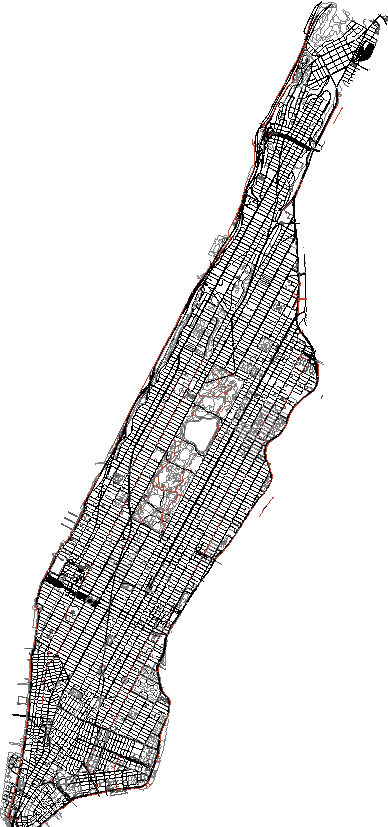}
    \caption{Manhattan road network.}
    \label{fig:Manhattan_network}
\end{figure}

In this experiment the target network is Manhattan (Fig.~\ref{fig:Manhattan_network}), which involves 3,971 TSCs, some complex intersections being managed by several TSCs, and 55,641 lanes. This network provides a testbed for evaluating the generalization and scaling properties of G-IG-RL approaches.

\smallskip
Training MARL based methods on that many intersections would imply massive time (running the corresponding simulations) and memory (storing and training one model per intersection) resources. Training S-IG-RL would enable the use of a constant number of parameters, but gathering the amount of experience required for RL on such a large road network would not be possible in a reasonable time with the simulator and computational resources that we use. Since it can leverage zero-shot transfer (i.e. no additional training) and keep the number of parameters constant, G-IG-RL, which is trained on a set of small networks as described earlier, is the only learned method usable in this large-scale context.  

For a given road network, G-IG-RL-L requires fixed computational time, while both the computational time and memory requirements of G-IG-RL-V increase linearly with the number of vehicles in the network.
 Considering the size of the experiment, even though we expect G-IG-RL-V to perform better than G-IG-RL-L considering the results of the previous experiment, we only evaluate the latter. 
 
Evaluating methods on this large network is more computationally demanding compared to Experiment 1. Every method is only evaluated on 5 instances of randomly generated traffic for a fixed duration of one hour and run using a single traffic regime which introduces a vehicle every second in the network. Compared with what G-IG-RL experienced during training, and what was evaluated in Experiment 1, adding a vehicle every second in a network as large as Manhattan leads to light traffic density which increases in certain areas as vehicles enter the network following asymmetric distributions based on a given random seed. Waiting for all trips to end in a large network would mean prolonging the duration of some simulations in an unreasonable way, considering that some trips might require a lot of time to be completed. For computational concerns, episodes end after an hour. We report aggregated results, but trips cannot be compared via pairing, as done in Experiment 1, without introducing bias.

 \subsubsection{Results}
\begin{figure}[htp]
    \centering
    \includegraphics[width=9cm, height=8cm]{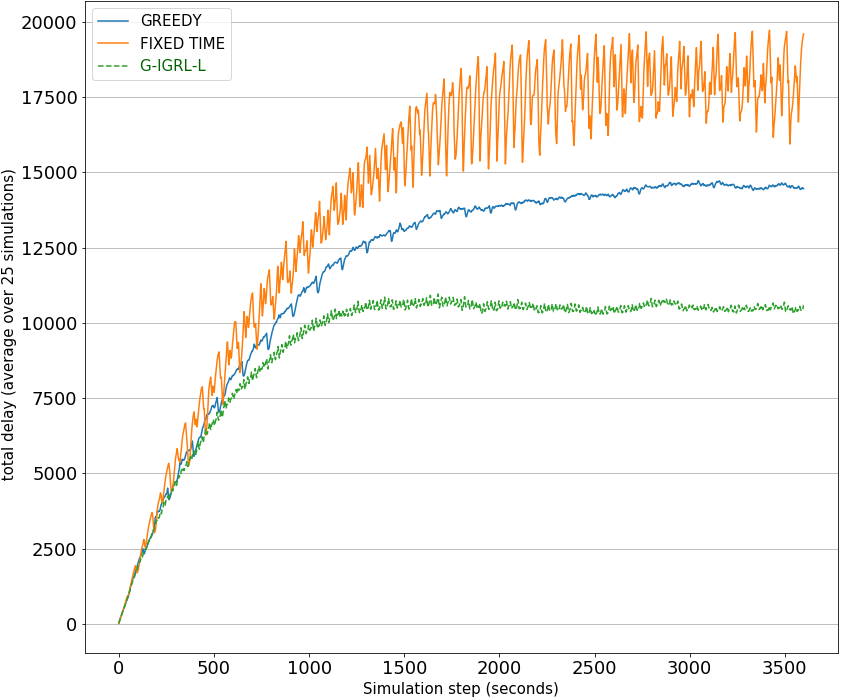}
    \caption{Total Delay Evolution: \textbf{Light} Traffic Regime | Manhattan Island.\\\hspace{\textwidth}As all simulations are running on the same road network, and the fixed time controls of many intersections are synchronized, the local delays are highly correlated across intersections and across simulations under this method, hence the periodic pattern.}
    \label{fig:step_delay_Manh}
\end{figure}
A first result is the bare fact that by using G-IG-RL-L, the 3,971 traffic signal controllers are operable, using a single GPU, and a model involving only 11,076 parameters. In addition, we can see that the total instantaneous delay, measured at every time step and reported in Fig.~\ref{fig:step_delay_Manh}, shows that G-IG-RL outperforms baselines which can scale to Manhattan, in terms of the slowest congestion increase and the lowest congestion-level steady-state. Furthermore, 
it confirms the ability of
IG-RL to generalize to different (here lighter) traffic regime than experienced in training. These results demonstrate that using synthetic randomly generated road networks for training is a viable and efficient way to learn policies that can transfer well to real road networks. This evaluation setting is similar in spirit to what is done in sim2real for robotics~\cite{sadeghi2018sim2real,akkaya2019solving}.

\section{Conclusion}
 We introduce IG-RL, a reinforcement-learning trained method that leverages the flexible computational graphs of GCNs and their inductive capabilities~\cite{hamilton2017inductive} to obtain a single set of parameters applicable to the control of a variety of road networks. 
 Experiments show that training using small-synthetic networks is enough to learn generalizable patterns, which in turn enables zero-shot transfer to new road networks, as well as new traffic distributions. Experiments also show that IG-RL outperforms MARL-IQL, as well as both dynamic heuristics and fixed-time baselines. Further, generalizability over architectures, which emerges from a generalist IG-RL (G-IG-RL), helps improve performance and generalization to new regimes of traffic. In addition, we showed that the flexibility of GCNs enables a flexible representation of the road network. Demand and structure can be represented and exploited in various ways, at their finest level of granularity, no matter the evolution of the number of entities nor their respective locations in the road network.

\paragraph{Future work}

IG-RL opens a path for the following future works: 1) The flexibility of GCNs could be key to the optimization of multi-modal transportation. We represent vehicles as nodes in IG-RL-V and we could experiment with pedestrian nodes, cyclist nodes, and many more, to ensure that traffic signal control accounts for all road-users instead of prioritizing cars. 
2) In this work, the reward function used for every MDP focuses on local queues (Equation~\ref{equation:reward}). In such a setting, coordination between MDPs is limited, and a shallow GCN is able to perform well since local information is sufficient. Evaluating whether the use of additional coordination mechanisms can further improve global performance could be an interesting avenue for future work. One way of addressing coordination, in the context of more global reward functions, would be to perform message passing on longer network distances by adding more layers to the GCN, or by adding recurrent mechanisms to it as done in~\cite{zhang2018gaan}. 3) The present work assumes data availability but some events (e.g. sensor failures) could generate various types of missing data. Evaluating IG-RL's robustness under several missing data mechanisms may constitute the object of future studies.   

\section{Supplementary Material}
\label{sec:supplementary_material}
\href{https://github.com/FXDevailly/IG-RL/blob/master/Supplementary_Material.pdf}{Supplementary material}, provided as a separate document, includes pseudo-code, a discussion on time complexity, and a complete version of Fig. \ref{fig:step_delay_L}.

\section{Acknowledgements}
We would like to thank the three anonymous reviewers for their interesting and constructive comments that lead to an improved version of this article. We would also like to thank Maxime Gasse for helpful discussions of deep learning on graphs which helped formulate GCN architectures included in this work.

\bibliographystyle{IEEEtran}
\bibliography{IEEEabrv,bibliography}

\end{document}